\documentclass[preprint,12pt,3p,twocolumn]{elsarticle}
\usepackage{amssymb}
\journal{Artificial Intelligence in Medicine}

\begin{document}
\begin{frontmatter}

\title{PEFT-MedAware: Large Language Model for Medical Awareness}
\author{Keivalya Pandya}
\ead{keivalyapandya2001@gmail.com}
\affiliation{organization={Birla Vishvakarma Mahavidyalaya},
             city={Vallabh Vidyanagar},
             country={India}}

\begin{abstract}
Chat models are capable of answering a wide range of questions, however, the accuracy of their responses is highly uncertain. In this research, we propose a specialized peft-MedAware model where we utilize parameter-efficient fine-tuning (PEFT) to enhance the Falcon-1b large language model on specialized MedQuAD data consisting of 16,407 medical QA pairs, leveraging only 0.44\% of its trainable parameters to enhance computational efficiency. The paper adopts data preprocessing and PEFT to optimize model performance, complemented by a BitsAndBytesConfig for efficient transformer training. The resulting model was capable of outperforming other LLMs in medical question-answering tasks in specific domains with greater accuracy utilizing limited computational resources making it suitable for deployment in resource-constrained environments. We propose further improvements through expanded datasets, larger models, and feedback mechanisms for sustained medical relevancy. Our work highlights the efficiency gains and specialized capabilities of PEFT in medical AI, outpacing standard models in precision without extensive resource demands. The proposed model and data are released for research purposes only. 
\end{abstract}


\end{frontmatter}


\section{Introduction}
\label{intro}
In recent years, we've witnessed a remarkable transformation in how we access information and tackle complex questions. This transformation has been driven by the incredible capabilities of chat models, which are AI-powered systems designed to respond to a vast array of queries. However, within the critical domain of healthcare, where precision and reliability are paramount, there exists a notable research gap. Existing models often struggle to bridge the gap between general information and specialized medical knowledge, as we will see in the coming sections. This research is driven by the pressing need to address this gap. While Large Language Models (LLMs) have the potential to revolutionize medical awareness, their effectiveness hinges on their ability to provide accurate and detailed information tailored to specific medical queries.

LLMs play a crucial role in advancing medical awareness. The medical field is a complex, ever-evolving domain with a constant influx of new research and discoveries. People often turn to the internet to find answers to their health-related questions. LLMs have the potential to revolutionize this information-seeking process by offering accurate, contextually relevant, and up-to-date medical information.

We propose peft-MedAware which stands for Parameter-Efficient Fine-Tuning-based \cite{liu2022fewshot} Medical Awareness model. This model is a Falcon one billion \cite{penedo2023refinedweb} parameter model fine-tuned over a specialized dataset with several training parameters that are optimized for quicker training time and using limited resources. We have used a comparatively lighter model, keeping information retrieval \cite{lewis2021retrievalaugmented} efficiency in mind. The primary focus of peft-MedAware is to bridge the gap between general health information and specialized medical knowledge by providing precise, reliable, and contextually relevant responses to medical queries.

\section{Literature Review}
\label{lit rev}
In this section, we discuss the related work and improvements to the existing approaches. There has been significant research in the field of natural language processing (NLP) aimed at improving the performance of chat models in answering questions accurately.

The research Xu, C. et al (2023) \cite{xu2023baize} proposes a new technique called Self-Distill with Feedback to improve the performance of the Baize chat model. The authors also introduce a new evaluation platform for open-domain dialogue systems and compare their approach with existing methods. The resulting Baize model demonstrates good performance in multi-turn dialogues and is released for research purposes. However, the possibility of generating factually incorrect or biased outputs, as well as the need for caution when using the model for commercial purposes due to legal considerations draw limitations to the Baize model. Abacha, A. B. (2019) \cite{ben-abacha-etal-2019-overview} proposes a novel approach to question answering based on recognizing question entailment (RQE). The authors compare logistic regression and deep learning methods for RQE using different datasets, including textual inference, question similarity, and entailment in both open and clinical domains. On the downside, the research includes the need for a large collection of question-answer pairs for real-time QA systems and the potential impact of the definition of RQE on QA results. Ming Zhu (2020) \cite{zhu-etal-2020-question} et al., discuss the challenges of developing effective question-answering (QA) systems that can handle long multiple-span answers. The authors propose using attention mechanisms to identify relevant parts of the input passage and generate accurate answers. They also suggest using deep learning techniques such as recurrent neural networks (RNNs) and transformer architectures to handle longer contextual dependencies. The approach only works for questions that require long multiple-span answers, and may not generalize well to other types of questions, the dataset used in this study is relatively small, which may limit the generalizability of the results and it relies heavily on the quality of the input data, which may be difficult to obtain in practice. Jiangui Chen (2021) \cite{Chen_2021} et al., present a novel approach to address the challenges faced by traditional machine learning techniques when dealing with heterogeneous and distributed QA data. The authors propose a federated learning framework called FedMatch, which enables training a powerful QA model using decentralized and non-identically distributed data that multiple parties to collaboratively train a single QA model without sharing their private data, thus preserving privacy and addressing issues related to data ownership and control. It assumes that all participating nodes have access to the same set of questions, which may not always be the case in real-world scenarios. In situations where different nodes have access to different sets of questions, this approach may not be effective.

On the other hand, our proposed model peft-MedAware is fine-tuned on a relatively smaller LLM focuses on information retrieval from a wide variety of diseases, and can provide trusted information from authentic resources to prevent the spread of any misleading information. One of the key advantages of peft-MedAware is its ability to handle a wide range of medical queries. Whether it's a general health question or a specific inquiry about a rare medical condition, peft-MedAware can generate accurate and contextually relevant responses. This versatility is crucial in the field of healthcare, where the diversity of medical queries is vast and constantly evolving.

\begin{figure*}[ht]
\includegraphics[width=1\textwidth]{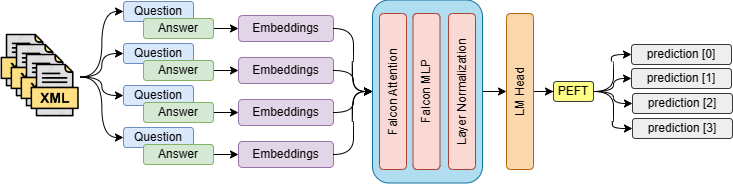}
\caption{peft-MedAware Pipeline}
\end{figure*}

\section{Dataset}
\label{data}
MedAware was trained on the MedQuAD \cite{Ben_Abacha_2019} dataset, a large-scale medical question-answering dataset of 47,457 QA pairs created from 12 NIH websites. The collection covers 37 question types (e.g. Treatment, Diagnosis, Side Effects) associated with diseases, drugs, and other medical entities such as tests. It is observed that out of all these QA pairs (as mentioned officially), more than half of them are left unanswered leaving us with 16,407 QA pairs along with the question type. The dataset is well-annotated, with each question-answer pair labeled with the type of evidence that supports the answer. Since each QA in MedQuAD is unique, we do not divide the dataset into train and test sets.
\subsection{Data Preprocessing}
This is a critical step to transform raw data into a format suitable for training machine learning models. Here, we present a Python function tailored for parsing the MedQuAD dataset, a medical question-answering dataset, and transforming it into a structured format suitable for further analysis and model training.

\section{Model Training}
\label{model}
\subsection{Parameter-Efficient Fine Tuning (PEFT)}
Training large language models requires millions, or even billions, of parameters. Although these parameters are necessary to capture complex patterns in a linguistic context, they pose computational, memory, and energy management challenges. Traditional optimization methods struggle to strike a balance between better performance and computational efficiency. PEFT on subtly optimizing the parameters of the model, ensuring the best use of computational resources. This method is especially important when dealing with large objects. \cite{sourab2023} By identifying and fine-tuning only the most influential parameters, the computational overhead is significantly reduced. This streamlined approach minimizes the time and resources required for training, making it feasible to experiment with various hyperparameters and configurations efficiently. The model that has been used contains over 700 million parameters out of which we utilize 3 million trainable parameters (that is, 0.44\% of overall parameters) to fine-tune based on our dataset.

\subsection{Quantization}
BitsAndBytesConfig \cite{bitsandbytes} configurations are essential components in the training pipeline of transformers. They dictate how data is processed, how weights are updated, and how the model learns from the provided dataset. This makes training large models feasible on diverse hardware configurations.

\begin{table*}[t]
    \centering
    \begin{tabular}{|c|p{10cm}|}
    \hline
         Human &  What are some of the symptoms of Cancer?\\
         \hline
         ChatGPT & Cancer is a group of diseases characterized by the uncontrolled growth and spread of abnormal cells. The symptoms of cancer can vary widely depending on the type of cancer and its stage. However, some common symptoms that might indicate the presence of cancer include:
        
        1. Fatigue: Persistent, unexplained fatigue that…
        
        2. Unexplained Weight Loss: Sudden and unexplained weight…
        
        3. Pain: Ongoing pain, such as headaches…
        
        4. Skin Changes: Yellowing of the skin…
        
        It's important to note that these symptoms can be caused by conditions other than cancer. However, if you experience any persistent or unusual symptoms, it's crucial to consult a healthcare professional for a proper evaluation and diagnosis. Early detection often significantly improves the chances of successful treatment for many types of cancer.\\
        \hline
        peft-MedAware & Signs and symptoms of cancer include new lumps or lumps that have gotten bigger or are not the same size as before. These lumps may be in the breast, lung, stomach, or other part of the body. They may also be in the head, neck, or bones.\\
        \hline
    \end{tabular}
    \caption{Comparison between ChatGPT and proposed model \textit{(not cherry-picked)}}
    \label{tab:table1}
\end{table*}

\begin{table*}[ht]
    \centering
    \begin{tabular}{|c|p{10cm}|}
    \hline
         Human &  I have a swollen knee and it makes some clicking sound when I bend it. What’s wrong with me?\\
         \hline
         ChatGPT & I'm not a doctor, but there are several potential causes for a swollen knee with clicking sounds when you bend it. Some possible reasons include:

        1. Arthritis: Osteoarthritis or rheumatoid arthritis can cause…
        
        2. Meniscus Tear: The meniscus is a cartilage in the knee that…
        
        3. Ligament Injury: Injuries to the ligaments in the knee, such…
        
        4. Bursitis: Bursitis occurs when the bursae, small…
        
        Given the variety of potential causes, it's crucial to have a healthcare professional evaluate your symptoms.\\
        \hline
        Baize-healthcare & Clicking or popping sound in your knee could be due to a variety of reasons like overuse, strain, and inflammation. You may want to see a doctor to get an accurate diagnosis.\\
        \hline
        peft-MedAware & You may have a torn meniscus. Meniscus is a soft, rubbery tissue that cushions the knee joint. Meniscus injuries are common in athletes. They can occur during sports or in everyday activities.\\
        \hline
    \end{tabular}
    \caption{Comparison between ChatGPT, Baize-healthcare, and proposed model \textit{(cherry-picked from \cite{xu2023baize})}}
    \label{tab:table2}
\end{table*}

\section{Model Setting}
\label{setting}
While configuring our parameters we have loaded our model using 4-bit precision which reduces the memory footprint and enables the training process to efficiently handle our chosen model and large datasets. \cite{quantization} This reduction in precision does not compromise the ability of the model to learn complex patterns. Moreover, we have used the double quantization technique where the 4-bit precision is employed in a more nuanced manner, which ensures that even with reduced bit precision, the model can capture intricate features of the data, enhancing its learning capabilities while maintaining computational efficiency. The 16-bit binary floating-point representation for computational operations strikes a balance between computational efficiency and numerical accuracy.

The model conserves memory and computational resources by using lower-bit precision and optimized quantization techniques. This reduction in requirements enables the training of more extensive models or facilitates the simultaneous training of multiple models without overwhelming the system resources especially while accessing Google Colab T4 Runtime for a limited amount of time. This also leads to faster training times. With reduced computational complexity, the model can iterate through epochs more swiftly.

\section{Conclusion}
\label{conclusion}
As technology advances, the specialized knowledge of platforms is redefining the landscape of medical consultations. The PEFT based model fine-tuning reduces the computational overhead significantly. By conserving memory, reducing computational complexity, and optimizing resources, our research benefits from faster training times and improved efficiency. We have observed that the PEFT approach, along with optimized quantization techniques makes the training process easier and efficient. While general AI models provide valuable information, they lack the depth required for accurate diagnoses. ChatGPT, despite its vast knowledge base, provides a generic overview of potential causes, lacking the specificity needed for an accurate diagnosis. Its response, while informative, lacks the depth required for medical accuracy. Baize-healthcare's response is vague and non-committal. It acknowledges the symptoms but does not delve into specific conditions. While it encourages consulting a doctor, it does not provide any concrete information, leaving the user uncertain about potential causes. On the contrary, our proposed model peft-MedAware's response demonstrates a remarkable level of precision and expertise. This ability to offer a specific diagnosis stems from its specialized focus on medical scenarios.

The model has been deployed to HuggingFace as https://huggingface.co/keivalya/peft-MedAware and this space carries all the updated information regarding the model.

\section{Limitations}
\label{limit}
\subsection{Safety and Access Control}
Users must keep their data safe and private when feeding it into the system. Be mindful especially with sensitive details to prevent any unauthorized access or data leaks. Plus, access to the model and its results should only be given to authorized people. To make sure this happens, using the right access controls and authentication methods is important. These measures help keep the system in check and make sure it's used responsibly.
\subsection{Ethical Considerations}
Language models, despite their impressive capabilities, can unintentionally inherit biases from the data they're trained on. To counter this, it's crucial to stay vigilant and work continuously to promote inclusivity and fairness. Additionally, we must consider the well-being of the users by protecting them from misinformation and ensuring that the model's responses don't encourage harmful activities. Providing clear disclaimers is essential, especially when the model's limitations could influence users' decisions.

\section{Acknowledgment}
\label{acknowledgement}
The author does not mean to disqualify or criticize any other research or researcher but would like to thank models such as Baize-healthcare \cite{xu2023baize} for undergoing such complex research and inspiring others to pursue it further.

\section{Discussion}
\label{discuss}
The model can be further improved in several ways. Primarily, to generalize it over more diseases either from scraping authorized websites or making larger datasets. Moreover, a larger model that can tune more parameters can be used to enhance model accuracy and adaptability. Incorporating a feedback loop from medical professionals and real-world users could provide valuable insights, helping refine the model's responses and ensuring it remains up-to-date with evolving medical knowledge.

\bibliographystyle{elsarticle-num} 
 \bibliography{bibfile}

\end{document}